\documentclass[letterpaper, conference, 10 pt]{IEEEtran}
\usepackage{times}

\usepackage{multicol}
\usepackage[usenames,dvipsnames,table]{xcolor}
\usepackage[bookmarks=true]{hyperref}
\hypersetup{
    colorlinks=true,
    linkcolor=orange,
    filecolor=magenta,      
    urlcolor=orange,
    citecolor=orange,
}
\usepackage{pdfx}
\usepackage{booktabs}
\usepackage{xspace}
\usepackage{amsmath}
\usepackage{caption}
\usepackage[capitalize]{cleveref}
\usepackage{graphicx}
\usepackage{inconsolata}
\usepackage[all]{hypcap}
\usepackage{multirow}

\usepackage{amsmath} 
\usepackage{amssymb}  

\vspace{0.15in}
\title{\LARGE \bf
\vspace{0.15in}
Efficiently Generating Expressive Quadruped \\ Behaviors via Language-Guided Preference Learning
\vspace{-0.15in}
}
\vspace{-0.15in}

\author{Jaden Clark$^{1}$ and Joey Hejna$^{1}$ and Dorsa Sadigh$^{1}$
\vspace{-0.15in}
\vspace{-0.15in}
\thanks{*This work was not supported by any organization}
\thanks{$^{1}$Jaden Clark, Joey Hejna, and Dorsa Sadigh are with the department of Computer Science,
        Stanford University, Stanford, CA 94305, USA, 
        {\tt\small jvclark@stanford.edu, jhejna@stanford.edu, dorsa@stanford.edu}}%
}

\begin{document}
\renewcommand{\arraystretch}{0.8}

\maketitle
\vspace{-0.2in}
\thispagestyle{empty}
\pagestyle{empty}

\vspace{-0.15in}
\begin{figure*}[h]
  \centering
  \includegraphics[width=0.97\linewidth]{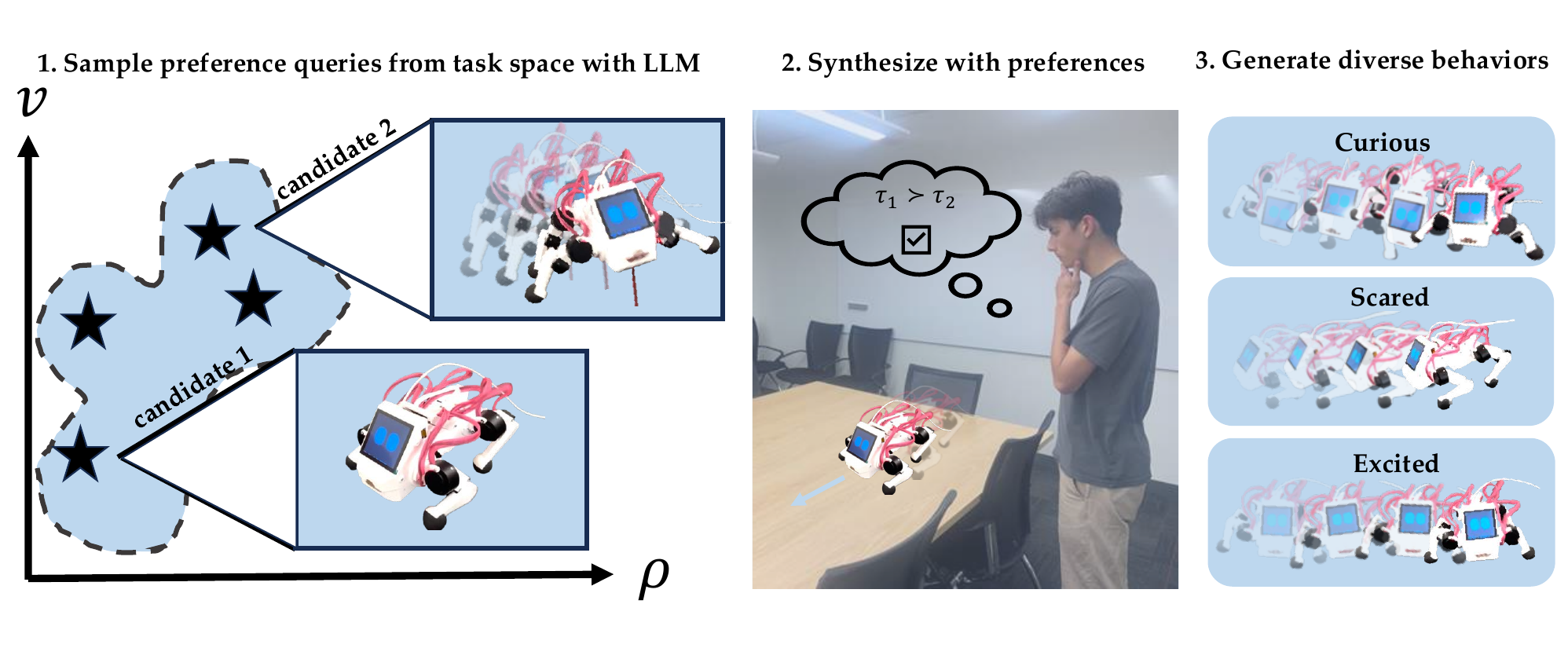}
  \caption{Our approach, LGPL, leverages LLMs to generate high quality candidate behaviors (1) for preference learning (2). This enables efficient design of diverse and accurate behaviors, even by non-expert users (3).}
  \label{fig:method}
\end{figure*}
\vspace{-0.15in}

\begin{abstract}

Expressive robotic behavior is essential for the widespread acceptance of robots in social environments. Recent advancements in learned legged locomotion controllers have enabled more dynamic and versatile robot behaviors. However, determining the optimal behavior for interactions with different users across varied scenarios remains a challenge. Current methods either rely on natural language input, which is efficient but low-resolution, or learn from human preferences, which, although high-resolution, is sample inefficient. This paper introduces a novel approach that leverages priors generated by pre-trained LLMs alongside the precision of preference learning. Our method, termed Language-Guided Preference Learning (LGPL), uses LLMs to generate initial behavior samples, which are then refined through preference-based feedback to learn behaviors that closely align with human expectations. Our core insight is that LLMs can guide the sampling process for preference learning, leading to a substantial improvement in sample efficiency. We demonstrate that LGPL can quickly learn accurate and expressive behaviors with as few as four queries,  outperforming both purely language-parameterized models and traditional preference learning approaches. Website with videos: \href{https://lgpl-gaits.github.io/}{this http url}.

\end{abstract}

\section{INTRODUCTION}

Recent advancements in reinforcement learning (RL) for quadrupeds have enabled agile locomotion across many complex terrains through diverse and tuneable gaits \cite{margolis2023walk, kumar2021rma, lee2020learning}. However, choosing the appropriate gait remains an open question, and selecting the incorrect gait may inadvertently generate dangerous or erratic movements that fail to align with human expectations \cite{hadfield2017inverse}. In the case of quadrupeds, the fastest gait might not always be suitable. For example, a more agile gait might optimize speed, but could pose dangers in social or crowded environments. This demonstrates the need for a principled method to generate human-aligned reward and objective functions.

Selecting the correct objective is particularly challenging when there is a need to quickly adapt. In the real world robots will interact with diverse humans to complete various desired tasks.  In the case of a social robot, different users desiring to interact with a robot may prefer it to move differently based on their age, comfort level with robots, or task at hand. For example, a younger child may prefer a slower and calmer gait whereas an older user may desire more expressive, rapid locomotion to accomplish a task. Here, the robot must quickly adjust its behavior to suit the expectations and needs of different users before they lose interest in the robot.

The need for user controlled gaits that can rapidly adapt to a user's preferences has led to a number of recent works that use a natural language due to its intuitiveness and ability to adapt to diverse environments  \cite{kwon2023reward, yu2023language, tang2023saytap, ma2023eureka}. However, natural language based approaches are fundamentally low-resolution, and it can be challenging for humans to describe the nuanced differences between robot gaits. For example, most humans are not experts on different quadruped locomotion styles. They may not know how to differentiate pacing and trotting, or what different step frequencies look like. So, it can be challenging to specify how a robot should walk without observing and providing specific feedback for existing gaits. On the other hand, preference learning allows users to compare different model outputs to provide high-resolution feedback. This makes it effective for refining robotic behaviors to align with human expectations where direct specification of complex reward functions is impractical \cite{macglashan2017interactive, christiano2017deep, lee2021pebble, hejna2023few}. Unfortunately, though learning from preferences allows for greater specificity it comes at a cost — determining a users intent from comparisons alone often requires hundreds \cite{hejna2023few} or thousands \cite{lee2021pebble} of queries for the most efficient approaches.  Querying a human's feedback hundreds of times to tune the quadruped's behavior may be tiresome for the user, and discourage them from adjusting the behavior as needed.

This work presents a hybrid approach that combines the versatility of language models with the precision of preference learning for quadruped locomotion. Our core insight is that although language is a coarse mechanism of feedback, it can provide high-quality candidate solutions which vastly speed up the preference learning process.

Specifically, our approach Language-Guided Preference Learning leverages large language models (LLMs) to generate candidate reward functions for a quadruped using in-context prompting. Then, humans provide feedback by ranking the LLM-generated behaviors. This feedback is then used to learn a final reward function, enabling the rapid customization of robot behaviors to user-specific requirements. This approach not only accommodates the broad behavioral adaptability facilitated by LLMs but also capitalizes on the capability of preference learning to estimate a user's true intent when it may be difficult to exactly specify with natural language. Through simulations and user studies, LGPL generates a diverse array of human-aligned behaviors with as few as four queries - generating behaviors that have 53\% lower L2 loss than preference learning and 62\% lower L2 loss than LLM parameterization based on MSE loss with respect to a ground truth reward (LGPL MSE was 0.223, preference learning MSE was 0.455, and LLM MSE was 0.821). Furthermore, we found that users preferred the behaviors from LGPL 76\% of the time when generating and tuning expressive behaviors over competing methodologies.

\section{Related Work}

Our work builds on existing research in quadruped locomotion, LLMs in robotics, and preference learning. Here, we review the areas most pertinent to our approach.

\begin{figure*}[h]
  \centering
  \vspace{0.05in}
  \includegraphics[width=0.93\linewidth]{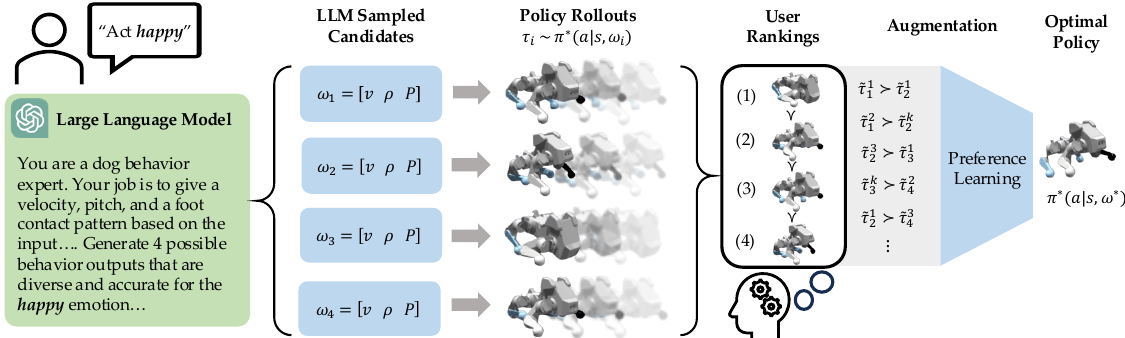}
  \caption{An overview of the LGPL method. A LLM is provided with in-context examples of task parameterizations $\omega$ that correspond to different quadruped gates, described in language. Given a new desired behavior, the LLM produces candidate parameterizations, which are used to rollout the policy. A user then ranks these candidates, which are eventually used for preference learning to discover the optimal $\omega$.}
  \label{fig:method}
  \vspace{-0.2in}
\end{figure*}

\noindent\textbf{Quadruped Locomotion.} A large body of work has shown that it is possible to train agile locomotion policies via approaches such as reinforcement learning (RL) capable of generating many different gaits \cite{caluwaerts2023barkour, fu2021minimizing, margolis2023walk, kumar2021rma, lee2020learning, yang2022fast}. However, such policies can be challenging to train without a sophisticated learning curriculum \cite{lee2020learning}, and gaits can be difficult to specify without manually encoding their reward parameters \cite{fu2021minimizing}. Several works propose methods for real time specification of gaits \cite{margolis2023walk, tang2023saytap}. However, these works are limited as they either do not consider how a non-expert user can tune the commands for the robot, or do not allow refinement of behaviors beyond low-resolution language feedback. On the other hand, LGPL provides an efficient means for both expert and non-expert users to rapidly tune desired gaits.

\noindent\textbf{Specifying Robot Behavior with LLMs.} A large body of recent work has focused on using LLMs to generate robot behaviors \cite{liang2023code, brohan2023can, yu2023language, mahadevan2024generative, thumm2024text2interaction, ma2023eureka}. Some works focus on using LLMs to generate long-horizon sequences of actions for robots \cite{brohan2023can, liang2023code}, and more recent work focuses on improving existing LLMs for robot-specific tasks \cite{liang2024learning}. Most relevant to our work is literature that explore generating reward functions directly from language instructions or corrections \cite{kwon2023reward, hu2023language, yu2023language}. Notably, several works use in-context learning to prompt an LLM to define reward parameters for a variety of robot tasks from high level commands such as ``sit'' or ``stand up'' \cite{yu2023language}, to more complex gait contact patterns \cite{tang2023saytap}. 
While this approach effectively bridges high-level behavioral commands with lower-level locomotion, it does not focus on refining these behaviors based on user feedback, which is critical in personalized interactions. 
 Furthermore, many robot behaviors such as differences in locomotion styles cannot be easily specified using language by a non-expert user \cite{casper2023open}. This brings us to consider how to adaptively learn reward functions from human feedback.

\noindent\textbf{Preference Learning.} 
Due to the challenges in manually specifying rewards for dynamical systems, several techniques have been developed to learn reward functions from human feedback. These methodologies encompass learning from 
physical corrections \cite{li2021learning}, scalar feedback \cite{knox2009interactively}, rankings \cite{myers2022learning, brown2019extrapolating}, and pairwise comparisons \cite{hejna2023few, lee2021pebble,sadigh2017active, ziegler1909fine}. 
These works often actively generate the next query to ask from the user in order to efficiently learn the reward function. Several methods employ preference-based methods to capture complex objectives but often suffer from sample inefficiency \cite{lee2021pebble, hejna2023few}. 
To improve sample efficiency, several works develop active learning strategies, some with an LLM in the loop \cite{peng2024preference, mahmud2024maple}, but often struggle in high-dimensional spaces \cite{sadigh2017active, biyik2018batch}. Moreover, active preference learning is still slow, requiring many samples and multiple feedback iterations. On the other hand real-time systems, like those for quadruped locomotion, necessitate simple and fast feedback mechanisms. Thus, instead of focusing on how to improve feedback over multiple iterations as in active learning, we instead focus on how to generate near-optimal initial reward candidates. Concurrent to our work \cite{yu2024few} also use an LLM provided with in-context information to sample candidate reward functions, which are then evaluated by human users before being used as future in-context examples. In contrast, our work focuses on using preferences to elicit real-time behavior from a fixed multi-task policy.

In the next section, we show how we can use LLMs to do so efficiently for a real-time quadruped locomotion system. Here we focus on learning a reward vector that is initialized via parameters generated from an LLM given an initial language instruction. We then further adapt such reward functions to align with the non-expert user's preferences via standard preference learning techniques. 


\section{Method}
\label{sec:lgpl}

In this section, we first describe the general problem of quadruped behavior adaptation with real-time human feedback. Then, we describe our approach, Language-Guided Preference Learning (LGPL), which uses an LLM to generate candidate reward parameterizations as shown in \cref{fig:method}. These candidates are then used as queries for preference learning to further adapt the rewards via rankings from human feedback.

\subsection{Problem Statement}

Consider a Markov Decision Process (MDP) with state space $S$, action space $A$, and dynamics $P$ defined as $(S, A, P, r_{\omega})$, where the  $\omega \in \Omega$ serves as a task-specific parameter that defines the reward function $r_\omega$. We train a task-conditioned policy $\pi^*(a|s,\omega)$ via reinforcement learning (RL) to maximize the expected reward $r_{\omega}$ across all tasks $\omega \in \Omega$. It is assumed that the distribution $\omega$ is accessible during the training phase (which allows for the learning of the optimal task-conditioned policy). However, our approach deviates from standard goal-conditioned RL by positing that the ultimate desired task $\omega^*$ is unknown and must be inferred from human feedback instead of being sampled from a fixed distribution.

In this context, we define the few-shot preference-based RL challenge. Our objective is to identify the desired task $\omega^*$ from  human feedback, while minimizing the number of user interactions required. In the real world, providing human feedback can be an arduous process \cite{lee2021b}, far too slow to adapt behaviors in real-time. Thus, to enable real-time systems we focus specifically on the few-shot setting where we have access to only a very small number of interactions with the user. 

To efficiently learn the task $\omega$, we assume that the user provides rankings over a small number of trajectories $(\tau_1, \tau_2, \dots, \tau_n)$ of length $T$, where $n < 10$. Each trajectory can be written as $\tau_i = (s_{1,i}, a_{1,i}, s_{2,i}, a_{2,i}, ..., s_{T,i}, a_{T,i})$. 
Following prior work \cite{wilson2012bayesian}, we assume that human responses follow the Bradley Terry model \cite{bradley1952rank}. Thus, the probability of one trajectory being preferred over another within the ranking can be written as

\begin{equation}
\label{eq:pref}
\hspace{-1em}
 P[\tau_1 \succ \tau_2] = \frac{\exp\left(\sum_t r_{\omega^*}(s_{t,1}, a_{t,1})\right)}{\exp(\sum_t r_{\omega^*}(s_{t,1}, a_{t,1})) + \exp(\sum_t r_{\omega^*}(s_{t,2}, a_{t,2}))}     
\end{equation}
where \( \tau_1 \succ \tau_2 \) indicates that trajectory $\tau_1$ is preferred to trajectory $\tau_2$ and $r_{\omega^*}$ is the users intended reward function. The users' ranking of $n$ trajectories thus results in $\binom{n}{2}$ such pairwise comparisons.

A typical approach would elicit user preferences over trajectories \( \tau_1, \tau_2, \dots, \tau_n \) of length $T$ generated by randomly sampling from $\Omega$ to generate corresponding \( \omega_1, \omega_2, \dots, \omega_n \). These preferences would then be used to learn a reward model. However, when $n$ is small sampling only a few random $\omega$ to generate trajectories is likely to be uninformative as the space of possible tasks is too large to cover in only a handful of queries. As a result, prior works on few-shot preference learning often assume access to optimal demonstrations \cite{brown2020safe}. However, providing human demonstrations in our setting is unrealistic, as this amounts to knowing the desired task vector $\omega$ apriori.
Moreover, intelligent sampling and active learning approaches which often rely on model uncertainty are equally ineffective with only a few data points. Instead, we posit that low-quality priors (e.g. random) are the source of this inefficiency. Next, we detail how we can tap into the world knowledge of LLMs to remedy this problem by generating better candidate $\omega$.

\begin{figure*}[h]
  \centering
  \includegraphics[width=0.95\linewidth]{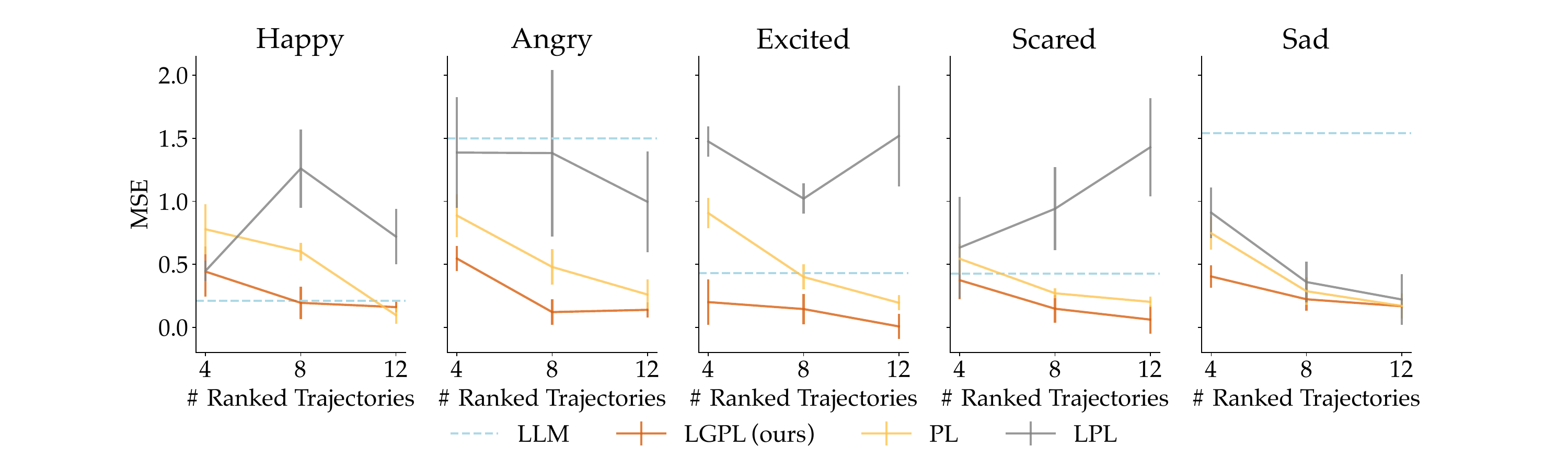}
  \vspace{-0.05in}
  \caption{Results of our simulation study to evaluate query efficiency. For each task, an expert-defined ground truth $\omega^*$ was chosen and the MSE between $\omega^*$ and the learned $\omega$ is plotted.}
  \label{fig:simresults}
\end{figure*}
\vspace{-0.1in}
\subsection{Language Parameterization}

The core insight of our method is that we can use LLMs to act as high-information priors. Specifically, we use an LLM to generate initial candidate tasks parameterizations $\omega$, which are subsequently evaluated based on the trajectories they produce. To enable the LLM to generate realistic quadruped behaviors, we prompt the model with pairs of task parameterizations $\omega$ and their corresponding language descriptions $l$ as provided by an expert user. We place these examples in the LLM context and ask the LLM to generate new candidate $\omega$ for user specified tasks in the form of the provided examples \cite{yu2023language}.  In particular, an LLM generates a series of candidate tasks \( \omega_1, \omega_2, \ldots, \omega_n \) as vectors based on the users language instruction. The LLM is deliberately prompted to generate task parameterizations that explore a range of possible behaviors, focusing on the key aspects of locomotion that are critical to the desired robot performance (see \cref{fig:method}). For example, when prompted to generate ``happy'' behavior, the LLM will generate reward vectors corresponding to trotting and bounding gaits at various different velocities and pitches. Critically, because of their ability to understand language and code, estimates generated by the LLM can be far closer to the intended behavior than random samples. Preferences over these candidates are subsequently far more informative for identifying $\omega^*$.

\subsection{Preference Learning}

Given the candidate task parameterizations generated by the LLM, we rollout the policy  $\pi^*(a|s, \omega)$ for each sampled candidate $(\omega_1, \omega_2, \ldots, \omega_n)$ yielding trajectories $( \tau_1, \tau_2, \dots, \tau_n)$ which can be visualized by a human user. After being shown all trajectories, the user ranks all $\omega_i$s, which in turn generates comparisons that can be used for preference learning. 


A na\"ive  approach would directly learn $\omega$ from the $\binom{n}{2}$ comparisons generated by the use rankings. However, in the few-shot setting where $n$ is small, this still results in only a handful of datapoints. To construct more training data, we assume that user preferences also hold over sub-segments of the trajectories \cite{park2022surf}. Using sub-segments of length $k$, each trajectory $\tau_i$ of length $T$ admits $T - k$ sub-segments $\Tilde{\tau}_i^1, \Tilde{\tau}_i^2, \dots, \Tilde{\tau}_i^k$ where $\Tilde{\tau}_i^j = (s_{j,i}, a_{j,i}, \dots, s_{j+k,i}, a_{j+k,i})$. In total, this results in $(T - k)^2 \binom{n}{2}$ comparisons that can be used for preference learning, which is substantially more than if attempting to directly learn $\omega$ from the original $n$ ranked values. Slightly overloading notation, we denote the modified dataset of all sub-segment comparisons as $\Tilde{D} = \{(\Tilde{\tau}_1, \Tilde{\tau}_2, y)\}$. where $y = \{1, 2\}$ specifies which of the two sub-segments is preferred.

To learn the user's intended task parameterization $\omega^*$ from the preference dataset $\Tilde{D}$, we minimize the binary cross entropy between the empirical preference distribution and the preference model implied by the learned parameterezation $\omega$. Succinctly, our objective is: 




\begin{align}
\label{eq:objective}
L_{pref}(\omega) = &-E_{(\Tilde{\tau_1}, \Tilde{\tau_2}, y) \sim \Tilde{D}} \left[ y(1) \log \left(P_\omega[\Tilde{\tau_1} \succ \Tilde{\tau_2}]\right) \right. \nonumber \\
&\left. + y(2) \log \left(1 - P_\omega[\Tilde{\tau_1} \succ \Tilde{\tau_2}]\right) \right]
\end{align}
where $P_\omega [\tau_1 \succ \tau_2]$ is the implied preference distribution from substituting the learned $\omega$ into $r_\omega(s,a)$ in \cref{eq:pref}. In other words, this is the standard binary cross entropy loss where the logits are determined by the cumulative reward estimates $r_\omega$ over a segment \( \Tilde{\tau} \). The goal is to maximize the probability of preferred preference segment — and consequently, the predicted reward values—of the preferred segment relative to the unpreferred one. By minimizing \cref{eq:objective} we obtain an estimate of $\omega^*$. Critically, at deployment time we do not need to retrain a policy. Instead, we just rollout the existing task conditioned policy $\pi^*(a|s, \omega)$ with $\omega^*$, facilitating real-time deployment. We call our approach Language-Guided Preference Learning (LGPL).

A requirement of LGPL is that the reward function $r_\omega$ is differentiable with respect to $\omega$. Fortunately, reward functions used for quadrupeds are often expressed as a sum of differentiable functions parameterized by features like velocity, orientation, and energy expenditure penalties. Thus, in LGPL we model $r_\omega$ as a sum of differentiable terms $r_\omega(s,a) = \sum_j \alpha_j \phi_j(s,a,\omega_j)$ where $\alpha_j$ are weights, $\phi_j(s,a,\omega_j)$ are parameter-specific differentiable reward functions, and $\omega_j$ is the $j$th element of the vector representation of the task.  This not only allows us to optimize $\omega$ using gradient descent, but also makes producing candidate $\omega$ with an LLM easy, as it just needs to output a vector. In the next section, we provide further details about how we use LGPL for a real-time quadruped system.

\section{Experiments}
\label{sec:experiments}

In this section, we seek to answer the following questions: 1) Is LGPL more sample efficient than existing preference learning approaches 2) Do humans prefer LGPL behaviors to behaviors from LLM and preference learning approaches? 3) Is LGPL fast enough to adapt to users in real time?

\begin{figure}[h]
  \centering
  \vspace{-0.15in}
  \includegraphics[width=0.975\linewidth]{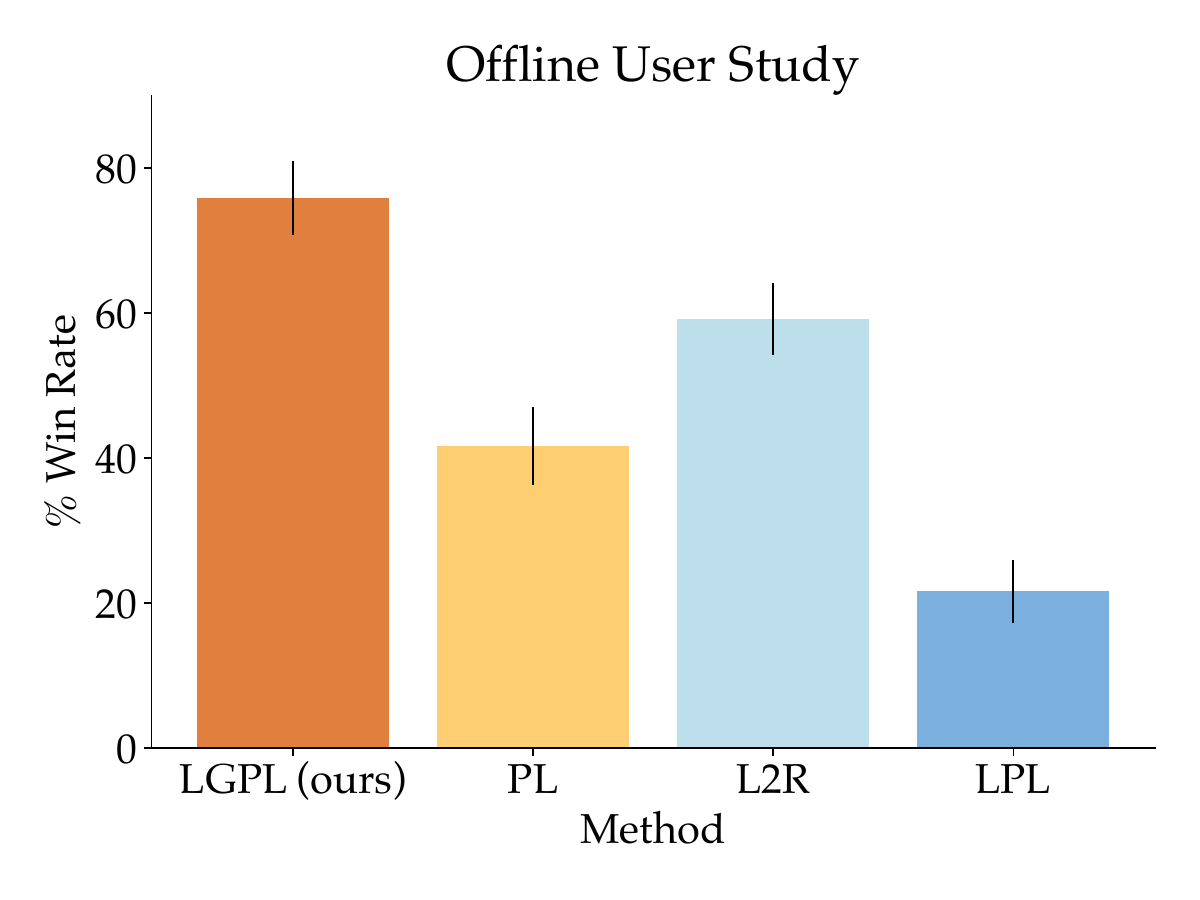}
  \vspace{-0.15in}
  \caption{Results for the offline user study. \% win rate signifies the percentage of time users preferred LGPL.}
  \label{fig:offline_study}
\end{figure}
\vspace{-0.2in}
\subsection{Implementation}
To test the efficiency of LGPL, we design a space of tasks $\Omega$ that is able to represent a broad set of behaviors. Specifically, we choose $\omega \in \mathbb{R}^5$ to be the desired velocity, desired pitch, and three essential primitives: trotting \( P_1 \), pacing \( P_2 \), and bounding \( P_3 \). We represent each gait by an indicator function where
\begin{equation*}
    \mathbf{1}_{\{gait=P_i\}}(gait) = 
\begin{cases} 
1 & \text{if the gait is } P_i, \\
0 & \text{otherwise}.
\end{cases}
\end{equation*}
In practice when learning $\omega$ as a continuous vector in $\mathbb{R}^5$ it may not strictly follow the form $[v, \rho, \mathbf{1}_{\{gait=P_1\}}, \mathbf{1}_{\{gait=P_2\}}, \mathbf{1}_{\{gait=P_3\}}]$ due to the indicator variables. For deployment we set $\mathbf{1}_{\{gait=P_i\}} = 1$ only for the maximum of the gaits and set the rest to zero. Our initial experiments found that this space of $\Omega$ was sufficient to express a broad variety of gaits desired by diverse users. We train a train an RL policy conditioned on $\omega$ to maximize $r_\omega(s,a) = \sum_j \alpha_j \phi_j(s,a,\omega_j)$, where $\phi_j$ are the aforementioned differentiable reward factors. For example, in our case at timestep $t$ $\phi_2(s,a,\omega_2) = \| \rho -\rho_t \|^2$ where $\rho_t$ is the pitch at $t$. For the different gaits, we set $\phi_j$ based on the contact pattern generator from prior work \cite{tang2023saytap}.
The task-conditioned policy is trained using the Nvidia Isaac Gym simulator \cite{makoviychuk2021isaac} with PPO \cite{schulman2017proximal}. The resulting policy $\pi^*(a|s,\omega)$, a feed-forward neural network, outputs desired positions for each motor joint. Inputs to this network include the robot's base angular velocities, the normalized gravity vector \(\mathbf{g} = [0, 0, -1]\) in the base frame, current joint positions and velocities, outputs from the previous timestep, and the targeted reward parameterization $\omega$. During training, $r_\omega$ was uniformly randomized to ensure the resulting policy is optimal across all tasks.

For deployment, we used a custom LLM prompt (defined in the project page) to generate a set amount of diverse task parameterizations.  In all experiments, we used GPT-4 as the LLM \cite{achiam2023gpt}. We then deploy our policies in the real world using the Pupper v3 quadruped. Pupper v3 has three joints per leg (hip, thigh, and calf joints), and is capable of diverse locomotion behaviors.

\subsection{Is LGPL more sample efficient than existing preference learning approaches?}

First, we assess how well LGPL can identify a wide variety of desired tasks with only a handful of preference queries. We compare LGPL against the following baselines:

\begin{itemize}
    \item Language to Reward (L2R): the LLM directly parameterizes the reward vector as done in \cite{yu2023language}.
    \item Preference Learning (PL): the human ranks queries randomly generated from the space of rewards the policy is trained on. These ranking are then used for preference learning \cite{brown2019extrapolating}.
    \item LLM ranked: After sampling rewards as in LGPL, we ask the LLM to re-rank its own generations and use those for preference learning
\end{itemize}

For all methods using language, we use chain-of-thought prompting to elicit stronger behavior specification \cite{wei2022chain}.

We consider 5 types of emotive behavior tasks in simulation -- "happy", "sad", "scared", "angry", and "excited" -- each corresponding to a unique $\omega^*$ provided by a human expert. For methods using preferences, we provide oracle preference labels by ranking trajectories according to the sum of rewards under the desired task, $\sum_t r_{\omega^*}(s_t,a_t)$. We evaluate methods on how well can predict the desired task vector by measuring the mean-squared error (MSE) with the ground truth $\omega^*$ across 5 seeds for 4, 8, and 12 ranked trajectories.


On average, we find that LGPL (depicted in orange) achieves far lower MSE with the ground truth task than standard preference learning particularly with fewer ranked trajectories, indicating that LGPL's use of an LLM prior can accelerate preference learning. Standard PL (depicted in yellow) often performs worse than simply querying the LLM when only given 4 ranked trajectories as it is unable to identify the correct task from uniform samples as seen in \cref{fig:simresults}. While simply using in-context learning performs competitively in some tasks (e.g. ``Happy''), it appears more sensitive to the language prompt, and further refinement via LPL only sometimes improves performance.

\subsection{Do LGPL Rewards Align with Human Expectations?}

A benefit of LGPL is its ability to be used for real-time interactions with users. To assess LGPL's ability to generate human-aligned behaviors, we conduct a user study in which participants observe videos of a quadruped executing behaviors generated by LGPL and each baseline method. 

We generate a trajectory for each behavior for all four methods. Each trajectory is split into 5 segments. Then we query users for preferences between trajectories for a given behavior, using 3 seeds. Participants also ranked each each trajectory on a Likert scale from 1-7 for how closely it matched the desired behavior. A total of 11 users conducted the study, and each rated all four methods for the 5 pre-determined tasks used in the previous section. Results for the user study are presented in \cref{fig:offline_study} and the top half of \cref{table:likert}. We found that LGPL was preferred over competing methods $75.83 \pm 5.14$\% of the time by users, indicating that LGPL is able to generate more distinguishable motions that are preferred by users. This stands in comparison to standard PL (in yellow), which only had a win rate of approximately 40\%. LLM methods L2R (light blue) and LPL (dark blue) are unable to performance as well as LGPL likely because language feedback alone is too coarse to shape emotive behavior as shown in \cref{table:likert}. 


\subsection{Can LGPL Adapt to Real-Time User Feedback?}

\begin{figure}[h]
  \centering
  \includegraphics[width=0.95\linewidth]{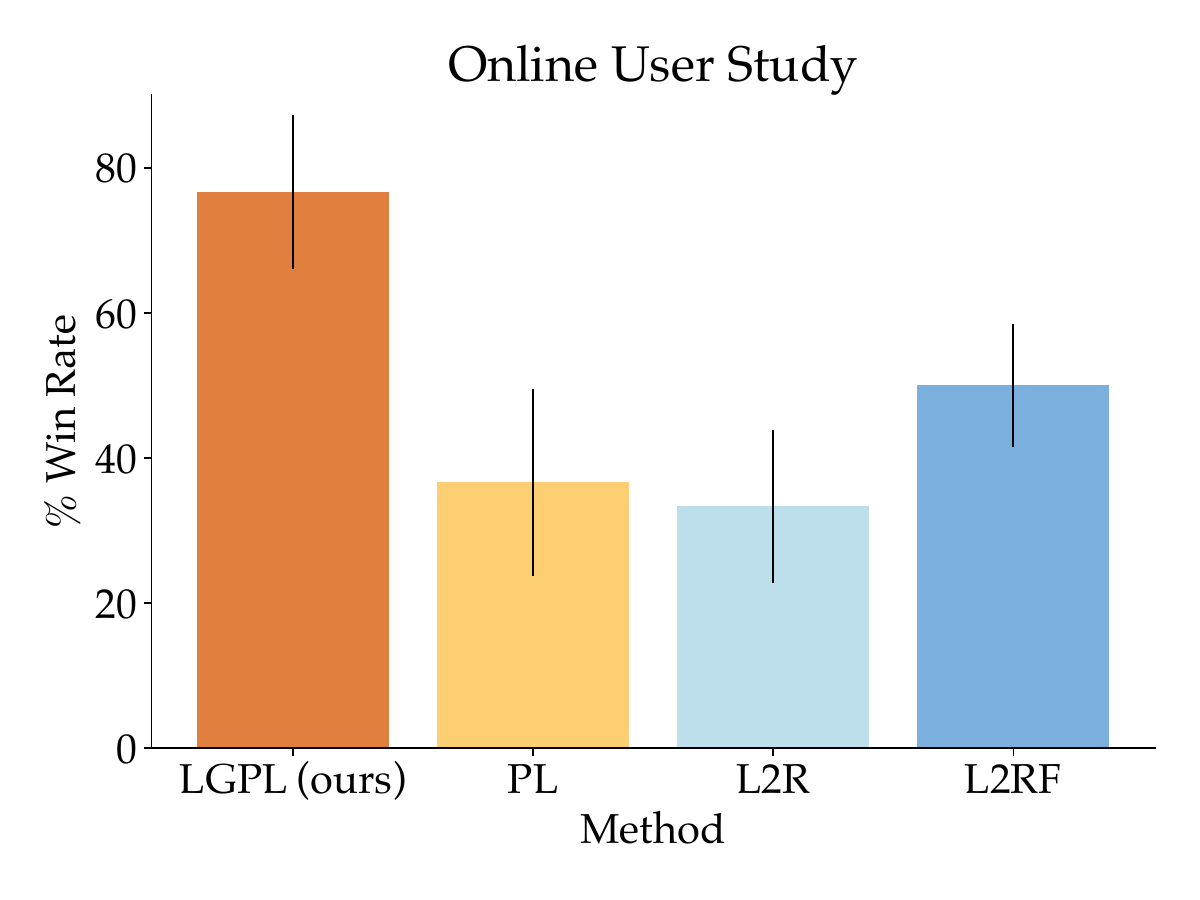}
  \vspace{-0.2in}
  \caption{Results for the feedback-driven user study. \% win rate signifies the percentage of time users preferred that method over all 3 other methods.}
  \label{fig:online_study}
\end{figure}

Finally, we seek to evaluate how efficiently LGPL can adapt to user feedback in real world. To do so, we conduct an additional user study in which all feedback is provided by real participants. Each user describes a desired behavior in language and then subsequently ranks 4 candidates for all methods that use preference learning (LGPL, PL) to estimate the intended task $\omega^*$. In addition to L2R, we consider an additional LLM-based baseline where we allow users to continue prompting the LLM to refine the behavior after observing rollouts of the L2R policy. We refer to this method as Language 2 Reward with Feedback (L2RF). Ultimately, users observe the final trajectories generated by all methods and assign each a Likert score from 1-7 and rank them according to how well they identify their intended behavior. The study consisted of 5 participants, and each participant evaluated all methods for two expressive tasks they described, such as fearful, nervous, playful, etc. The full study took approximately 20 minutes. Results are presented in \cref{fig:online_study} and the bottom half of \cref{table:likert}. We found that LGPL was preferred $76.67 \pm 10.59$\% of the compared  to baseline methods. Notably, in some cases providing language feedback to L2R often did not improve the new gait (L2RF), and L2R gaits before feedback were only preferred less 40\% of the time. Thus, when relatively few (4) queries are available, LGPL generates user-preferred behaviors for novel expressive tasks better than both preference learning and L2R. Furthermore, repeatedly querying semantic feedback from users is not necessarily a good alternative as providing accurate language feedback can be more onerous than rankings, and does not necessarily generate preferred trajectories.

\begin{table}[ht]
\centering
\small 
\caption{Likert Scores (1-7) for Active and Offline Conditions. Higher is better and the best scores are bolded.}
\begin{tabular}{lccccc}
\multirow{2}{*}{Condition} & PL & LGPL & LPL & L2R & L2RF \\
\cmidrule(lr){2-6} 
Offline Study & 4.44 & \textbf{5.46} & 3.90 & 4.33 & x    \\
STD & $ \pm 0.23$ & $\pm 0.21$ & $ \pm 0.30$ & $ \pm 0.21$ & x    \\
Active Study  & 4.0  & \textbf{5.5}  & x    & 3.9  & 4.2  \\
STD & $\pm 0.49$  & $\pm 0.41$  & x    & $\pm 0.59$  & $\pm 0.56$  \\
\bottomrule 
\end{tabular}
\label{table:likert}
\end{table}
\section{Discussion}
\label{sec:discussion}

\subsection{Conclusion}

In this paper, we introduce Language-Guided Preference Learning (LGPL), a new approach for efficiently generating  quadruped behaviors aligned with human preferences. By using LLMs to produce initial candidate reward functions and refining them through preference-based feedback, LGPL combines the versatility of language input with the precision of preference learning. Our experiments demonstrated that LGPL outperforms both traditional preference learning and purely language-parameterized models in terms of sample efficiency and accuracy, achieving desired behaviors with as few as four queries. Human evaluations confirmed that LGPL-generated behaviors closely match user expectations for expressions such as happiness, sadness, fear, and anger. These findings highlight the potential of LGPL to enable rapid and precise customization of robot behaviors in dynamic social environments.

\subsection{Limitations and Future Work}
Future work has the potential to scale LGPL to more complex real-time systems by overcoming its limitations.


\begin{itemize}
    \item \textbf{Differentiability.} LGPL requires differentiable reward functions, which potentially prohibits the expressiveness of possible tasks.
    \item \textbf{Chaining behaviors} While LGPL can efficiently identify individual tasks, real-world expressive behaviors may involve sequences of sequential tasks. \looseness=-1
    \item \textbf{Query difficulty} Behaviors produced by by preference learning algorithms can be hard for humans to discern. Future work could explicitly consider legibility when generating queries.
    \item \textbf{Dimensionality} LGPL works well with only a few queries, however increasing the dimension of the task vector $\omega$ may come at the cost of sample efficiency as both LLM generation and preference learning become harder. \looseness=-1
\end{itemize}

\textbf{Acknowledgments:} We thank Wenhao Yu for much insightful guidance. This work was supported by NSF \#2218760, and \#2132847, AFOSR YIP, DARPA YFA Grant \#W911NF2210214, and the DARPA FACT project.



\addtolength{\textheight}{-12cm}   






\bibliographystyle{IEEEtran}
\bibliography{references}

\end{document}